%% file: NAFBench_arxiv_new.tex
\documentclass[letterpaper]{article} 
\usepackage{aaai2027}  
\nocopyright
\usepackage[hyphens]{url}  
\usepackage{graphicx} 
\usepackage{natbib}  
\usepackage{caption} 
\usepackage{algorithm}
\usepackage{algorithmic}

\usepackage{newfloat}
\usepackage{listings}
\DeclareCaptionStyle{ruled}{labelfont=normalfont,labelsep=colon,strut=off} 
\floatstyle{ruled}
\newfloat{listing}{tb}{lst}{}
\floatname{listing}{Listing}

\usepackage{booktabs}
\usepackage{multirow}
\usepackage{fvextra}

\usepackage{xspace}
\usepackage{amssymb}
\usepackage{amsmath}
\usepackage{tikz}
\usetikzlibrary{arrows.meta,positioning,calc}

\definecolor{blFill}{HTML}{E6F1FB}\definecolor{blLine}{HTML}{185FA5}\definecolor{blTxt}{HTML}{0C447C}
\definecolor{tlFill}{HTML}{E1F5EE}\definecolor{tlLine}{HTML}{0F6E56}\definecolor{tlTxt}{HTML}{085041}
\definecolor{pkFill}{HTML}{FBEAF0}\definecolor{pkLine}{HTML}{993556}\definecolor{pkTxt}{HTML}{72243E}
\definecolor{amFill}{HTML}{FAEEDA}\definecolor{amLine}{HTML}{854F0B}\definecolor{amTxt}{HTML}{633806}
\definecolor{coFill}{HTML}{FAECE7}\definecolor{coLine}{HTML}{993C1D}\definecolor{coTxt}{HTML}{712B13}
\definecolor{nk}{HTML}{444441}

\title{Not What You Meant: Can LLMs Follow a Specified Negation Semantics?}
\author{
    Qiming Bao\textsuperscript{\rm 1},
    Agnieszka Mensfelt\textsuperscript{\rm 2},
    Michael J. Witbrock\textsuperscript{\rm 1},
    Kostas Stathis\textsuperscript{\rm 2}
}
\affiliations{
    \textsuperscript{\rm 1}University of Auckland, New Zealand\\
    \textsuperscript{\rm 2}Royal Holloway, University of London, United Kingdom\\
    qiming.bao@auckland.ac.nz,
    agnieszka.mensfelt@rhul.ac.uk,
    m.witbrock@auckland.ac.nz,
    kostas.stathis@rhul.ac.uk
}

\newcommand{\NAFBench}{\textsc{NAF-Bench}\xspace}

\newcommand{\code}[1]{\texttt{\small #1}}
\newcommand{\tid}[1]{{\small\texttt{#1}}}

\begin{document}

\maketitle

\begin{abstract}
Negation does not carry a uniform interpretation across domains. In legal, regulatory, and medical reasoning, the intended interpretation depends on the reading in force -- open- versus closed-world, two- versus three-valued, credulous versus skeptical. We study which reading of negation large language models adopt by default, and whether they can override that preference when a different reading is explicitly specified. To this end we introduce \NAFBench, a procedural generator of solver-certified instances spanning four semantic viewpoints: SLDNF, well-founded semantics (WFS), and credulous and skeptical reasoning under stable-model semantics. The generator emits ground normal logic programs with controlled depth, width, and cycle structure; each is solved under all four viewpoints with SWI-Prolog, a well-founded solver, and clingo, yielding up to four divergent labels; the programs are then verbalized into natural language under multiple framings and rule orderings that leave the answer invariant. The results expose a consistent gap. Across open-source models, following a specified negation semantics remains unsolved: the strongest scores $59$--$74\%$ across the four semantic viewpoints and the weakest $31$--$67\%$; all are order-sensitive on more than half of logically identical rule shufflings, while the two weaker models frequently overcommit on well-founded ``undefined.'' Two frontier models reach $100\%$ on the main fixed-complexity evaluation set, and a third (o4-mini) is near-perfect, falling only to $81\%$ on well-founded ``undefined.'' Delegating reasoning to a solver, fine-tuning on certified traces, or forcing an explicit three-valued verdict each partly closes the gap.
\end{abstract}


\begin{quote}
\small\itshape
``The dog did nothing in the night-time.''\\
``That was the curious incident.''

\hfill\normalfont\scriptsize
Arthur Conan Doyle, ``The Adventure of Silver Blaze'' (1892)
\end{quote}

\section{Introduction}

Reasoning from absent information is common in high-stakes decision making, and the intended meaning of ``not'' varies across settings. In law, 
the presumption of innocence offers an intuitive analogy for negation as failure: guilt must be established positively, and when every attempt to establish guilt fails, innocence is retained by default. By contrast, skeptical reasoning accepts a conclusion only if it holds in every possible account consistent with the available evidence.
In clinical and safety diagnostics, a fault may be inferred from the unavailability of its complementary fault, and mutual exclusion between the two yields genuinely undetermined states. In regulatory and benefits adjudication, an entitlement may hold under some admissible reading of the rulebook while failing under others. A single natural-language scenario can therefore license different conclusions depending on which negation semantics is in force. A system that supports such tasks should apply the semantics that has been specified, and should identify the cases in which a conclusion is undetermined under that semantics.

Large language models (LLMs), including reasoning-tuned models, are now deployed on tasks of this kind, and their logical behaviour has been characterised along several dimensions: high aggregate accuracy alongside variation under structural perturbations of the same problem \citep{bao2022pararuleplus, chen2024premise, mirzadeh2025gsm, bao2025robustness}, influence of prior beliefs on conclusions drawn from stated rules, described as logic inertia \citep{bao2025conflict}, and declining accuracy as the search space grows \citep{lin2025zebralogic,shojaee2025illusion}. Separately, across a range of task types, reliability is limited when a query admits no determinate answer and abstention is expected \citep{kirichenko2025}. 

Isolating negation reading and the propagation of undetermined conclusions requires a benchmark in which the semantics disagree and every answer is independently certified. Existing works cover adjacent ground. Lexical benchmarks evaluate the detection of negation in prose \citep{nguyen2023xnot360,so2026thunder}. Reasoning benchmarks fix a single, usually implicit, closed-world reading, so that observed failures remain consistent with several explanations: the semantics in force, problem complexity, and surface framing \citep{bao2022pararuleplus,lin2025zebralogic}. Recent work evaluates LLMs on answer-set programming tasks \citep{ren2026aspbench}, where evaluation targets task performance under a fixed semantics. The design we adopt treats negation reading as a controlled variable, supplies multiple answers per instance, and defines failure as not following the reading given in the prompt.

We introduce \NAFBench, a procedural generator of solver-certified instances built around this control. The generator emits ground normal logic programs with controlled depth, width, and cycle structure, and certifies each program under SLDNF, well-founded semantics (WFS), and stable-model semantics, recording both credulous and skeptical entailment using
SWI-Prolog, 
a well-founded solver, and clingo so that a single instance carries up to four divergent certified answers to a query (Figure~\ref{fig:running}). 
Two axes are varied independently -- nominal complexity (negation depth, width, cycle length) and label divergence (the certified four-viewpoint signature) -- and solver hardness per instance is co-registered, so observed failures can be attributed to a source.
Programs are verbalized under multiple framings, languages, and rule orders that leave the certified answer invariant, separating semantic-viewpoint competence from sensitivity to framing, order, and length, and permitting difficulty to be regressed onto structure with token count held fixed. We evaluate models in two conditions: with the intended negation semantics stated in the prompt, and with the semantics left unspecified.

We use six models in two panels: three frontier systems (\texttt{claude-sonnet-5}, \texttt{gpt-5.6-sol}, \texttt{o4-mini}) and three open-weight models in the $8$--$32$B range (\texttt{qwen2.5-coder-32b}, \texttt{deepseek-r1-32b}, \texttt{llama3-8b}). With the semantics left unspecified, the two frontier models converge on the same default reading of negation: assert a definite truth value when the negation cycle admits at least one internally consistent assignment of truth values, and answer cannot be determined when it varies or when no resolution exists. The two models agree on all 120 items in this condition, a pattern not reproduced by any constant-answer baseline. Once a semantics is named, results separate by the model panel. \texttt{claude-sonnet-5} and \texttt{gpt-5.6-sol} answer every instance correctly, and an inspection of their traces indicates the correct application of the semantics. \texttt{o4-mini} reaches $94.8\%$, its errors concentrated on the well-founded condition, where it correctly identifies the negation cycle as undefined but then collapses that value to false instead of propagating it. The open panel is substantially weaker and shows no consistent default reading, within or across models; the analysis of the traces points to different modes of failure. A translate-then-solve baseline and a small fine-tuning study show improvement in following the specified semantics.

Because instances are synthetic, the benchmark is contamination-resistant
and regenerable as models improve, at some cost to ecological validity -- which we
mitigate by grounding the difficulty axes in the structures that make real legal,
diagnostic, and regulatory rulebooks diverge.

\section{Preliminaries}

\NAFBench is restricted to normal logic programs: default negation only, with no classical or double negation. 
This restriction ensures that all semantic frameworks considered here are defined over the same program syntax, allowing differences in conclusions to be attributed to their treatment of default negation rather than to differences in expressive power.

Rules are directional and have the form
\texttt{h :- b\_1, ..., b\_m, not c\_1, ..., not c\_n}, where \code{not}
denotes default rather than classical negation; the program language contains
no biconditional connective.

Four readings of default negation are used throughout, and we fix the following labels for them.
\begin{description}
  \item[\textsc{SLDNF}] Negation-as-failure under the closed-world assumption gives an operational, order- and loop-sensitive reading in which failure to prove $p$ 
  is treated as evidence for
  $not\ p$ \citep{clark1978naf,shepherdson1984negation}.
  \item[\textsc{WFS}] The well-founded semantics is three-valued, assigning \emph{undefined} to atoms founded neither true nor false, which arises for atoms in loops over negation \citep{vangelder1991wfs}.
  \item[\textsc{cred.}] Stable-model (answer-set) semantics admits zero, one, or many models \citep{gelfond1988stable}; credulous entailment holds for atoms true in some model.
  \item[\textsc{skept.}] Skeptical entailment under the same semantics holds for atoms true in every model.
\end{description}

A wider literature on negation in logic programming lies outside the benchmark's scope and is left for future work: classical negation distinguishes explicit falsity from failure-to-prove \citep{gelfond1991classical}, negation can be treated as inconsistency \citep{gabbay1986negation}, constructive and intensional negation extend the treatment to variable-bearing goals and to explicit synthesis of negative information \citep{chan1988constructive,liu1999constructive,barbuti1987intensional}, and contradiction can be removed within \textsc{wfs} \citep{alferes1998contradiction}. The credulous/skeptical distinction corresponds to credulous and skeptical acceptance in abstract argumentation \citep{dung1995acceptability}, whose application to legal reasoning \citep{benchcapon2020before} supplies the legal and regulatory narratives used in verbalization.

\section{The \NAFBench{} Benchmarking Tool}
\label{sec:tool}

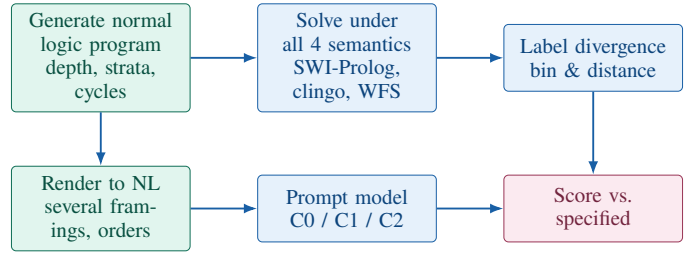
\begin{figure*}[t]
\centering
\resizebox{0.5\textwidth}{!}{%
\begin{tikzpicture}[
  node distance=6mm and 10mm,
  box/.style={draw=blLine,fill=blFill,text=blTxt,rounded corners=2pt,
              align=center,inner sep=4pt,font=\small,minimum height=10mm,
              text width=24mm},
  gen/.style={box,draw=tlLine,fill=tlFill,text=tlTxt},
  sco/.style={box,draw=pkLine,fill=pkFill,text=pkTxt},
  ar/.style={-{Latex[length=2mm]},blLine,thick}]
  \node[gen] (gp) {Generate normal logic program\\{\footnotesize depth, strata, cycles}};
  \node[box,right=of gp] (solve) {Solve under all 4 semantics\\{\footnotesize SWI-Prolog, clingo, WFS}};
  \node[box,right=of solve] (label) {Label divergence bin \& distance};
  \node[gen,below=8mm of gp] (nl) {Render to NL\\{\footnotesize several framings, orders}};
  \node[box,right=of nl] (prompt) {Prompt model\\{\footnotesize C0 / C1 / C2}};
  \node[sco,right=of prompt] (score) {Score vs.\ specified};
  \draw[ar] (gp) -- (solve);
  \draw[ar] (solve) -- (label);
  \draw[ar] (gp) -- (nl);
  \draw[ar] (nl) -- (prompt);
  \draw[ar] (prompt) -- (score);
  \draw[ar] (label) -- (score);
\end{tikzpicture}%
}
\caption{The \NAFBench{} generation-and-verification pipeline. A generated
program is (i) solved under all four semantic viewpoints to obtain certified labels and a
divergence bin, and (ii) rendered to natural language under several framings and
rule orders. The model is prompted under conditions C0 (no semantics named), C1
(semantics named, agreeing instance), and C2 (semantics named, divergent
instance); answers are scored against the certified label.
}
\label{fig:pipeline}
\end{figure*}

\NAFBench is a fully automatic pipeline (Figure~\ref{fig:pipeline}) with
solver-certified ground truth and no human annotation, so it can be regenerated
harder as models improve.

\paragraph{Negation readings} An instance is a pair $(P,q)$ where $P$ is a ground normal logic program
and $q$ an atom (the query). We study four semantics
$\mathcal{S}=\{\textsc{sldnf},\textsc{wfs},\textsc{stb}_\exists,\textsc{stb}_\forall\}$:
SLDNF, well-founded, and credulous ($\exists$) and skeptical
($\forall$) stable. For each $s\in\mathcal{S}$ a solver-certified labelling
$\lambda_s(P,q)\in\mathcal{L}_s$ returns the answer, with semantics-dependent
label spaces
\[
\begin{aligned}
  \mathcal{L}_{\text{\textsc{sldnf}}}
    &= \{\top,\bot,\circlearrowleft\},\quad
  \mathcal{L}_{\text{\textsc{wfs}}}
    = \{\top,\bot,u\}, \\
  \mathcal{L}_{\text{\textsc{stb}}_{\exists}}
    &= \mathcal{L}_{\text{\textsc{stb}}_{\forall}}
    = \{\top,\bot\},
\end{aligned}
\]
where $\circlearrowleft$ is SLDNF non-termination and $u$ is WFS undefined. The four coincide on acyclic programs; divergence requires a cycle, and cycles through negation are what \NAFBench targets.

Figure~\ref{fig:running} shows an example with divergent labels. Consider the loop $b\!\to\!c\!\to\!d\!\to\!b$, an
odd cycle through negation. \emph{Stable models}: a set $M$ is stable iff
it equals the least model of the Gelfond--Lifschitz reduct $P^M$. Suppose $b\in
M$; then \code{b :- not c} forces $c\notin M$; since \code{c :- not d} is $c$'s
only rule, keeping $c$ out of the least model of the reduct requires that rule
to be deleted, i.e.\ $d\in M$; and $d$'s support via \code{d :- not b} forces
$b\notin M$ -- a contradiction; the
symmetric attempt $b\notin M$ fails likewise.
No set is stable, so the program has zero stable models. Hence the
credulous query ($\exists M.\,a\in M$) is false, while the skeptical query
($\forall M.\,a\in M$) is vacuously true. \emph{Well-founded}: the
alternating fixpoint grounds no literal of the odd loop either way
($\Gamma(\emptyset)=\{a,b,c,d\}$, $\Gamma(\{a,b,c,d\})=\emptyset$), so $b$, and
therefore $a$, is \emph{undefined}. \emph{SLDNF}: resolving \code{a} unfolds into
the loop and does not terminate. 
Even and odd cycles behave differently: an even loop such as \code{e :- not f}, \code{f :- not e} has two stable models ($\{e\}$ and $\{f\}$), so a query on one side is credulously true but skeptically false, whereas the odd loop above admits none.

\begin{figure}[t]
\centering
\small
\setlength{\tabcolsep}{4pt}
\fbox{\parbox{0.9\columnwidth}{
\ttfamily
a :- b.\\ b :- not c.\\\ c :- not d.\\ d :- not b. \\[2pt]
\rmfamily\normalfont query: \ttfamily a\rmfamily?
}}\\[4pt]
\begin{tabular}{lll}
\toprule
Semantics & Output & Answer on \code{a} \\
\midrule
 SLDNF (operational) & 2-valued $+\;\circlearrowleft$ & \emph{non-termination} ($\circlearrowleft$) \\
Well-founded        & 3-valued & \emph{undefined} ($u$) \\
Stable, credulous ($\exists$) & 2-valued & No \\
Stable, skeptical ($\forall$) & 2-valued & Yes (vacuous) \\
\bottomrule
\end{tabular}
\caption{
A sample program and query with its value under each of the four semantics.
}
\label{fig:running}
\end{figure}

\paragraph{Generation.}
The generator emits ground normal logic programs with controlled negation depth
(nested \code{not} chain length), width (number of shared subgoals feeding the
query), and cycle structure (stratified, even, or odd loops through negation).
Depth and width are decoupled from cycle length, so structural complexity and the
phase boundary where semantics diverge can be varied independently.

\paragraph{Certification and divergence bins.}
Every generated program is solved under all four semantic viewpoints -- clingo
\citep{gebser2019clingo} for stable-model reasoning, a well-founded solver for
WFS, and SWI-Prolog \citep{wielemaker2012swipl} for SLDNF with an operating-system
timeout as a loop signal -- and is assigned to one of four divergence bins
by its certified four-viewpoints signature
$(\textsc{stb}_\exists,\textsc{stb}_\forall,\textsc{wfs},\textsc{sldnf})$:
\textbf{control} $(\top,\top,\top,\top)$, no cycle; \textbf{even-one-sided}
$(\top,\bot,u,\circlearrowleft)$; \textbf{odd} $(\bot,\top,u,\circlearrowleft)$
(no stable model, so skeptical is vacuously true and credulous false); and
\textbf{even-both-sided} $(\top,\top,u,\circlearrowleft)$. Bins are defined by signature, not by syntax, so instances where the semantics happen to agree are reassigned to control or rejected; conversely a control instance is one certified to have a constant signature, which entails that the query's relevant subprogram is acyclic.

\paragraph{Per-cell distinct-program variation.}
To obtain a population of items per cell rather than one canonical program,
we generate, at fixed depth and width, many structurally distinct but
gold-preserving programs by varying: the chain positions at which the query
depends on the cycle core and on the width block; the number of aggregator
predicates over a fixed set of shared subgoals (each subgoal retained by $\ge 2$
parents); the distribution of supporting facts at a fixed total; an always-true
guard literal on the cycle rules; and the rule order. Each candidate is
re-certified and kept only if its four-label signature matches the bin, and
isomorphic duplicates are removed by a canonical key, so a cell contains genuinely
different programs sharing the same certified answers
(Algorithm~\ref{alg:gen}).

\begin{algorithm}[t]
\caption{Certified per-cell generation}
\label{alg:gen}
\begin{algorithmic}[1]
\STATE \textbf{Input:} depth $d$, width $w$, target bin $b$, cycle length $k$, count $N$
\STATE $\mathit{kept}\gets\emptyset$;\ \ $\mathit{seen}\gets\emptyset$
\WHILE{$|\mathit{kept}| < N$}
  \STATE $\alpha\gets$ sample axes (attach points, aggregator count, fact
         distribution, guard literal, rule order)
  \STATE $P\gets \textsc{Assemble}(d,w,k,\alpha)$
  \STATE $\ell\gets \textsc{CertifyFull}(P,q)$ \COMMENT{clingo, WFS solver, SWI-Prolog}
  \STATE $\kappa\gets \textsc{CanonicalKey}(P)$
  \IF{$\textsc{Signature}(\ell)=b$ \AND $\kappa\notin\mathit{seen}$}
     \STATE add $P$ to $\mathit{kept}$;\ \ add $\kappa$ to $\mathit{seen}$;\ \ record $\ell$ and solver metrics
  \ENDIF
\ENDWHILE
\STATE \textbf{return} $\mathit{kept}$
\end{algorithmic}
\end{algorithm}

\paragraph{Verbalization, framing, and ordering (invariants).}
Programs are rendered rule-by-rule using either an abstract proposition-level template (e.g., ``proposition $x$ is true if proposition $y$ is not true'') or controlled narrative templates (e.g., ``\ldots{} if \ldots'' and the mutual-exclusion phrasing ``\ldots{} if and only if \ldots{} does not''). Thus, structural differences remain visible in the rendered rules rather than being hidden by a fixed scenario, and a leakage check ensures that the wording does not reveal the certified answer. We evaluate three nuisance transformations that preserve the underlying program and its certified labels:
\emph{framing} (alternative narrative vocabularies for the same program),
\emph{language} (matched English and Chinese renderings), and
\emph{ordering} (rule/fact permutations).
Order-invariance holds by definition for WFS and stable semantics and also holds for $\lambda_{\textsc{sldnf}}$ on the generated instances used here: the controls are acyclic and terminate under every tested rule ordering, whereas SLDNF evaluation of each divergent instance reaches a negative cycle and neither succeeds nor finitely fails under any ordering used in E4.
Any change in a model's answer across orders is therefore pure order-sensitivity, a property we measure directly.

\paragraph{Structural axes and solver effort.}
The generator sets two structural axes; solver effort is recorded per instance
alongside the labels.
\begin{itemize}
  \item \textbf{Nominal complexity} (varied): negation depth, width, and cycle
    length.
  \item \textbf{Label divergence} (varied): the certified signature
    $(\textsc{stb}_\exists,\textsc{stb}_\forall,\textsc{wfs},\textsc{sldnf})$,
    i.e.\ the divergence bin. Controls are the constant signatures, where every
    semantics yields the same answer; the divergent bins are where the choice of
    semantics determines it.
  \item \textbf{Solver effort} (recorded): choice points and conflicts (clingo),
    fixpoint iterations (WFS). Each semantic viewpoint invokes its own solver, so these are
    comparable within a semantics; on $\circlearrowleft$ instances we record the
    loop budget in place of SLDNF derivation counts, which there reflect the
    budget itself.
\end{itemize}
The two structural axes are independent: divergence follows from cycle parity and isolation
alone, so program size varies freely at a fixed signature, and the signature
varies freely at a fixed size, except that the odd bin requires an odd cycle and so differs from the even bins by one rule. Solver effort is recorded per instance and per
semantics, giving each result a measure of the work an exact algorithm expends.

\paragraph{Protocol}
Each item is presented under three conditions, in independent contexts so that no answer is contaminated by having already seen the program: \textbf{C0}, no semantics named, which elicits the model's default reading; \textbf{C1}, a named semantics on a control instance, where all four readings agree; and \textbf{C2}, the same named semantics on a divergent instance matched to that control in depth and width. C1 therefore isolates instruction-following from commitment to a particular reading of negation -- a correct answer there requires no choice among semantics -- while C2 forces the choice, and the C1-to-C2 drop is the quantity of interest. C0 answers are scored against all four certified labels, which identifies the semantic viewpoint with which a model's default reading agrees, or that it agrees with none.

\section{Experiments}

\paragraph{Setup.}
Our main evaluation set fixes a single cell (depth $8$, width $4$; cycle length $4$ for
even bins, $3$ for the odd bin) and, since the certified answer is invariant to
verbalization, uses one rule-level rendering. It comprises
$4~\text{bins}\times 30~\text{distinct programs}\times 5~\text{conditions}=600$
prompts over $120$ distinct programs (the four semantic viewpoints plus a no-instruction
baseline). We evaluate an open-source panel served locally -- Qwen2.5-coder~32B,
Llama3-8B, and DeepSeek-R1~32B -- at temperature $0$, single pass, together with
three frontier models (o4-mini, Claude Sonnet 5, GPT-5.6 Sol) on the identical
set (the frontier reasoning models do not expose a temperature control). All
$95\%$ confidence intervals are bootstrap intervals clustered over the
$120$ distinct programs rather than over prompts, so they reflect
program-to-program variation. Each answer is
scored against the solver-certified label under the requested semantics;
the no-instruction condition supplies each model's unprompted answer, used both for the agreement diagnostic below and to identify which semantics, if any, best matches that answer. Reasoning models that conclude in their own format (e.g.\
$\backslash$\code{boxed}$\{\cdot\}$) are read with an extractor that maps the
free-form conclusion to the label space, validated to agree with the strict
\code{ANSWER: X} parser wherever the latter succeeds.
Beyond answer accuracy we also audit the traces. For each response, we use a regex-based parser to extract the truth values it states for two items: the atom through which the negation cycle enters the positive chain, and the query. We compare these against the solver's certification: a trace is \emph{sound} if both verdicts are correct and \emph{contradicted} if either disagrees; traces that commit to no verdict are excluded. All reported rates are therefore approximate lower bounds (see appendix).

\paragraph{Instances, prompts, and trials.}
An \emph{instance} is a program--query pair $(P,q)$ with its four certified labels;
a \emph{prompt} renders it under a chosen semantics, framing, and rule order; a
\emph{trial} is one scored response. One program yields many prompts, and each
experiment varies a single axis with the program held fixed: E1--E2 use the $120$
main evaluation programs, E3 varies size ($74$ programs), E4 varies \emph{only} order
(Table~\ref{tab:order}) or framing, and the mitigations (E6) use a $44$-program
structurally-diverse WFS set (delegation and verify) plus a smaller
preliminary fine-tuning study.

\paragraph{Implementation.}
The open-source models are served locally with ollama (v0.5.7) on its
OpenAI-compatible endpoint, on a node with two NVIDIA A100-SXM4 80\,GB GPUs, an
AMD EPYC 7713 CPU, and 2\,TB RAM under CentOS Stream 8 (Linux 4.18). Certification
uses clingo 5.8, SWI-Prolog 9.3.5 (with a 2.5\,s OS-level timeout to detect
 SLDNF loops), and a Python well-founded solver; analysis uses SciPy 1.13.
Group-level significance is assessed with the two-sided Fisher exact test on
control-vs-divergent counts and confirmed by a Mann--Whitney $U$ test over
per-program correctness; size-moderator claims use a program-clustered bootstrap.
Generation is seeded per variant, and model decoding is deterministic
(temperature $0$), so runs are reproducible from the code and generated sets,
available at \url{https://github.com/14H034160212/NAFBench}. The leaderboard is available at \url{https://huggingface.co/spaces/qbao775/naf-bench-leaderboard}.

\begin{table}[tb]
\centering
\small
\begin{tabular}{lccccc}
\toprule
Model & cred. & skept. & WFS & SLDNF & cor. \\
\midrule
\multicolumn{6}{l}{\emph{Frontier}} \\
Claude Sonnet 5   & 100 & 100 & 100 & 100 & n/a \\
GPT-5.6 Sol       & 100 & 100 & 100 & 100 & n/a \\
o4-mini           & 100 & 98 & 81 & 100 & \phantom{0}4 \\
\midrule
\multicolumn{6}{l}{\emph{Open-source}} \\
Qwen2.5-coder 32B & 72 & 61 & 74 & 59 & 55 \\
Llama3-8B         & 62 & 67 & 36 & 31 & 69 \\
DeepSeek-R1 32B   & 72 & 64 & 31 & 43 & 47 \\
\bottomrule
\end{tabular}

\caption{Semantic-following accuracy (\%) per semantics and the default-correspondence
rate, main evaluation set. ``cred.''/``skept.'' are credulous/skeptical stable.
``cor.'' is a default-correspondence measure: of a
model's incorrect answers, the percentage that coincide with its own
no-instruction default reading (lower is better; n/a when a model makes no
errors). 95\% CIs (clustered over the 120 programs) span roughly $\pm 8$--$11$
points on accuracy for the open panel.}
\label{tab:main}
\end{table}

\paragraph{Overview.}
We report six experiments, each self-contained (shared protocol above; full
per-cell breakdowns in the appendix):
\begin{itemize}\setlength{\itemsep}{1pt}
  \item \textbf{E1} -- semantic-following accuracy per model and semantics (Table~\ref{tab:main});
  \item \textbf{E2} -- where the failure concentrates, by divergence bin (Table~\ref{tab:perbin});
  \item \textbf{E3} -- whether program size or the named semantics drives difficulty (Table~\ref{tab:size});
  \item \textbf{E4} -- robustness to rule order and surface framing (Table~\ref{tab:order});
  \item \textbf{E5} -- Performance on ``undefined'' (Table~\ref{tab:overcommit});
  \item \textbf{E6} -- three mitigations (Tables~\ref{tab:t2s}--\ref{tab:sft}).
\end{itemize}

\paragraph{E1: Semantic following remains challenging below the frontier.} Table~\ref{tab:main} shows a capability split. On the main fixed-complexity evaluation set, two frontier models -- Claude
Sonnet 5 and GPT-5.6 Sol -- score $100\%$ across all four semantic viewpoints, and o4-mini is near-perfect ($98$--$100\%$ on credulous, skeptical, and SLDNF), dropping only on well-founded \emph{undefined} ($81\%$). The open panel is far lower: the strongest (Qwen2.5-coder~32B) reaches only $59$--$74\%$ and the weakest (Llama3-8B) $31$--$64\%$. Accuracy is moreover an upper bound on competence here, because on the odd bin an odd cycle admits no stable model, so the skeptical answer is vacuously \emph{yes} -- the same letter produced by plain forward-chaining that never notices the cycle at all. Two open models are credited on $23/30$ and $25/30$ skeptical odd-cycle items and fall to $3/30$ and $2/30$ on the same thirty programs under the credulous reading (appendix), where that coincidence disappears; this collapse is measured on answers alone and assumes nothing about their traces. A trace audit points the same way with a weaker guarantee: among the traces its extractor can adjudicate, $14$--$38\%$ of the open panel's answers over adjudicable traces state a verdict that contradicts certification, against none of the frontier's $724$.

\paragraph{E2: The failure concentrates on divergent bins.}
Table~\ref{tab:perbin} breaks accuracy down by divergence bin (semantics named).
On the \emph{control} bin, where all four semantics agree so the default is
correct by construction, the two weaker models are strong (Llama3-8B $78\%$,
DeepSeek-R1 $81\%$); on the three \emph{divergent} bins they fall to $30$--$52\%$,
a drop of up to $\sim\!50$ points with \emph{no} change in program size. Qwen2.5-coder,
by contrast, is comparatively uniform ($62$--$72\%$) across bins. The gap between
a model's control accuracy and its divergent accuracy is itself a correspondence
signature: it is exactly the region where the default answer stops being right
that accuracy falls away. The control-vs-divergent gap is highly significant for
the two weaker models (Fisher exact $p<10^{-12}$ for both Llama3-8B and
DeepSeek-R1) but not for Qwen2.5-coder ($p=0.74$), the one model that does not
collapse.

\begin{table}[tb]
\centering
\small
\begin{tabular}{lcccc}
\toprule
Model & control & even-1 & odd & even-2 \\
\midrule
Qwen2.5-coder 32B & 65 & 67 & 62 & 72 \\
Llama3-8B & 78 & 30 & 37 & 51 \\
DeepSeek-R1 32B   & 81 & 38 & 52 & 35 \\
\bottomrule
\end{tabular}
\caption{Accuracy (\%) by divergence bin, semantics named (excludes the
no-instruction baseline). ``control'' = all four semantic viewpoints agree; ``even-1''
even-one-sided, ``even-2'' even-both-sided. Weaker models are strong on control
and collapse on divergent bins despite identical program size. Each bin has
$30$ programs $\times\,4$ semantics ($120$ prompts, minus a few unparsed).}

\label{tab:perbin}
\end{table}

\paragraph{E3: Difficulty is driven by semantics, not by size.}
On a larger depth$\times$width sweep -- $222$ prompts over $74$ distinct programs
(up to three per cell), all three open models -- accuracy is flat and non-monotone
along both axes (Table~\ref{tab:size}): no model improves or degrades
systematically as the program grows. A standardized OLS fit of per-item correctness on depth, width, and divergence bin -- on a separate sweep that spans all four bins (360 prompts; depths $\{2,8,16\}$, widths $\{0,4,8\}$) -- gives negligible size coefficients
($z(\text{depth})=-0.024$, $z(\text{width})=-0.006$) against dominant bin
coefficients ($-0.42$ to $-0.63$), and a bootstrap CI on
$|\hat\beta_{\text{width}}|-|\hat\beta_{\text{depth}}|$ includes $0$: neither size
axis is a significant moderator once the divergence bin is controlled.

\begin{table}[tb]
\centering
\small
\setlength{\tabcolsep}{4.5pt}
\begin{tabular}{lccccc}
\toprule
 & \multicolumn{5}{c}{negation depth (accuracy \%)} \\
\cmidrule(l){2-6}
Model & $d{=}0$ & $2$ & $4$ & $6$ & $8$ \\
\midrule
Qwen2.5-coder 32B & 71 & 67 & 67 & 64 & 64 \\
Llama3-8B         & 44 & 43 & 45 & 44 & 51 \\
DeepSeek-R1 32B   & 42 & 61 & 68 & 41 & 51 \\
\midrule
 & \multicolumn{5}{c}{shared-subgoal width (accuracy \%)} \\
\cmidrule(l){2-6}
Model & $w{=}0$ & $2$ & $4$ & $6$ & $8$ \\
\midrule
Qwen2.5-coder 32B & 69 & 73 & 62 & 62 & 68 \\
Llama3-8B         & 38 & 59 & 44 & 37 & 49 \\
DeepSeek-R1 32B   & 62 & 50 & 56 & 38 & 59 \\
\bottomrule
\end{tabular}
\caption{Marginal accuracy along the two nominal-complexity axes (depth sweep
averaged over width, and vice versa) on the larger grid ($222$ prompts, $74$
programs, even-one-sided bin): each of the $25$ depth$\times$width cells contributes up to three distinct programs $\times$ three conditions (credulous/skeptical/WFS). 
}

\label{tab:size}
\end{table}

\paragraph{E4: Models are sensitive to order- and framing-invariant changes.}
Two nuisance transforms leave the certified answer unchanged. \emph{Rule order:}
on $160$ (program, condition) groups rendered in four rule orders each, the three
open models change their answer across orders on a majority of groups
($54$--$67\%$, Table~\ref{tab:order}), whereas the frontier models are far
steadier ($0$--$15\%$) -- the same capability split as in accuracy.
\emph{Surface framing:} on $32$ groups each rendered under three vocabularies
(escalation / sensor-alarm / audit-override), the open models' answers flip across
framings on $41\%$ (Qwen2.5-coder), $41\%$ (Llama3-8B), and $23\%$ (DeepSeek-R1) of
groups, despite identical logic. The open panel is thus far from the invariance the
task demands, while the frontier models largely achieve it.

\begin{table}[tb]
\centering
\small
\begin{tabular}{lccc}
\toprule
Model & accuracy (\%) & flip (\%) & cov. \\
\midrule
\multicolumn{4}{l}{\emph{Frontier}} \\
Claude Sonnet 5   & 99.7 & \phantom{0}1 & 640/640 \\
GPT-5.6 Sol       & 100  & \phantom{0}0 & 640/640 \\
o4-mini           & 94.8 & 15 & 640/640 \\
\midrule
\multicolumn{4}{l}{\emph{Open-source}} \\
Qwen2.5-coder 32B & 68.1 & 54 & 640/640 \\
Llama3-8B         & 48.3 & 64 & 636/640 \\
DeepSeek-R1 32B   & 55.0 & 67 & 556/640 \\
\bottomrule
\end{tabular}
\caption{Rule-order robustness: the main evaluation programs re-rendered in four
logically-equivalent rule orders ($640$ prompts $=$ $160$ (program, condition)
groups $\times 4$). Accuracy is over all $640$ prompts; the \emph{order-flip rate}
is the fraction of groups whose answer is not constant across the four orders
(group counted if $\ge2$ orders parse); ``cov.'' is parsed prompts.
}
\label{tab:order}
\end{table}

\paragraph{E5: Weaker models underperform on ``undefined''.}
On well-founded items whose certified answer is \emph{undefined} (gold ``cannot
be determined''), the two weaker models return a definite yes/no on most of them
(Table~\ref{tab:overcommit}) -- forcing a two-valued verdict where the semantics
licenses none. Only Qwen2.5-coder abstains correctly ($91\%$). For Llama3-8B this is a
genuine two-valued collapse, visible in the traces: on a WFS-undefined item it
forward-chains through the negation loop as if it were an ordinary rule chain,
concluding ``\emph{\ldots{} so proposition q is true \ldots{} ANSWER: A}'' -- it
never entertains that the loop leaves the query ungrounded, exactly the
closed-world default the semantics was meant to override. DeepSeek-R1's low WFS
score ($31\%$) is only partly this collapse: a trace audit finds $80\%$ of
its WFS-undefined traces reach an ``undefined'' conclusion but only $26\%$ render
it as the choice ``C'' (the rest self-override or state it in prose a keyword
extractor mis-scores). Its reasoning competence thus exceeds its answer
accuracy -- a scoring caveat for free-form reasoning models, detailed in the
appendix.

\begin{table}[tb]
\centering
\small
\setlength{\tabcolsep}{5pt}
\begin{tabular}{lcc}
\toprule
Model & definite (A/B) & correct ``undef.'' \\
\midrule
Qwen2.5-coder 32B & 9\%  & 91\% \\
Llama3-8B         & 74\% & 26\% \\
DeepSeek-R1 32B   & 71\% & 29\% \\
\bottomrule
\end{tabular}
\caption{Behaviour on the main evaluation WFS-condition items whose certified answer is \emph{undefined} ($\approx 85$ parsed per model; same programs as E1). ``definite'' is the share answered A/B.
}
\label{tab:overcommit}
\end{table}

\paragraph{E6, mitigation 1: delegate the semantics to a solver.}
Asking the model only to \emph{autoformalize} the scenario into a logic program
(translate-then-solve) and letting the solver apply the semantics separates reading a rulebook
from applying it. On a larger, structurally-diverse set of $44$ WFS programs
(even cycles $k\in\{2,4,6\}$, conjunctive/disjunctive query support, and
controls), delegation lifts every open model far above its direct accuracy
(Table~\ref{tab:t2s}); the residual gap tracks translation fidelity -- each model's
accuracy is capped by how many of the $44$ programs it renders into a
parseable program ($44/44$, $42/44$, $38/44$). The results therefore indicate that applying the specified semantics is a
major bottleneck, although translation fidelity still limits performance.

\begin{table}[tb]
\centering
\small
\setlength{\tabcolsep}{6pt}
\begin{tabular}{lcc}
\toprule
Model & direct WFS & $+$solver (translate-then-solve) \\
\midrule
Qwen2.5-coder 32B & 59 & 95 \\
Llama3-8B         & 49 & 74 \\
DeepSeek-R1 32B   & 62 & 89 \\
\bottomrule
\end{tabular}
\caption{Translate-then-solve on $44$ structurally-diverse WFS programs (accuracy
\% of parsed answers).  
Per-model nulls, strict accuracy, and the verify-scaffold breakdown are
in the appendix.}

\label{tab:t2s}
\end{table}

\paragraph{Mitigation 2: fine-tune on solver-certified reasoning.}
A small LoRA fine-tune on solver-certified chains-of-thought recovers
near-perfect WFS accuracy on held-out items (Table~\ref{tab:sft}), and divergent
accuracy reaches $100\%$ on the fine-tuned models --evidence that the certifier can generate its own
repair data. Transfer is the caveat: the same adapter nearly saturates its trained
\emph{narrative} framing ($18/44\to40/44=91\%$, Table~\ref{tab:sft}) but does
\emph{not} transfer to an \emph{unseen abstract} framing of the same items
($30/44\to28/44=64\%$, no gain; appendix), so training must span framings and
augment over logically-equivalent transforms \citep{bao2024amrlda}.

\begin{table}[tb]
\centering
\small
\begin{tabular}{lccc}
\toprule
Model & base & +SFT & +SFT+DPO \\
\midrule
gemma-3-4b-it & 41 & 91 & --- \\
qwen2.5-7b    & 59 & 93 & 93 \\
\bottomrule
\end{tabular}
\caption{WFS accuracy (\%) after LoRA fine-tuning on solver-certified reasoning,
held-out 44-item set; the qwen $+$SFT cell is the mean of $3$ seeds ($91$--$96\%$).
}
\label{tab:sft}
\end{table}

\paragraph{Mitigation 3: verification scaffold, and language robustness.}
A prompt-only ``verify-before-answer'' scaffold (state each atom's value before
committing) improves \emph{every} open model on the same $44$-program WFS set -- %
Qwen2.5-coder $59\!\to\!82\%$, DeepSeek-R1 $62\!\to\!69\%$, Llama3-8B
$49\!\to\!62\%$ (parsed) -- so part of the reversion is a recoverable commitment
failure rather than a missing capability. The failure is also
language-robust -- the pattern persists when the programs are re-rendered in Chinese. Together the three mitigations
localize the gap to applying default negation and show it is at least
partly repairable, by delegation, targeted training, or making the three-valued
distinction explicit.

\section{Related Work}

\paragraph{Negation in NLP and LLMs.}
A separate line studies whether models detect orunderstand
negation at the sentence level. \citet{nguyen2023xnot360} find GPTs only modestly
proficient at recognizing when one sentence negates another (xNot360), and
Thunder-NUBench contrasts negation against contradiction and paraphrase for
sentence-level understanding \citep{so2026thunder}. This competence -- spotting
negation in prose -- is orthogonal to ours: we ask whether a model will
apply a specified formal negation semantics to structured rules whose
answer legitimately depends on which semantics is named.

\paragraph{Logical-reasoning benchmarks for LLMs.}
Rule-based reasoning benchmarks such as ProofWriter
\citep{tafjord2021proofwriter} and PARARULE-Plus \citep{bao2022pararuleplus}
render synthetic logic programs into language, reading-comprehension
benchmarks such as ReClor \citep{yu2020reclor} test logical inference over prose,
and abductive benchmarks test explanation of unexpected observations
\citep{young2022abductionrules}; all fix a single (implicit, closed-world)
reading. Structure-perturbation studies show that high scores hide
brittleness \citep{bao2025robustness}, and complexity-scaling studies
\citep{lin2025zebralogic,shojaee2025illusion} -- show accuracy collapsing as raw problem size grows.

\NAFBench is complementary and distinct: it varies nominal complexity (negation depth, width, cycle length) and the negation semantic viewpoint. Framing, language, and rule order are varied as certified-label-preserving nuisance transformations, so perturbation robustness is measured as well.

\paragraph{LLMs with logic solvers and autoformalization.}
A growing body of work couples LLMs to symbolic solvers, either translating
natural language to ASP/Prolog or using solvers to assist question answering, and
benchmarks the resulting pipelines
\citep{wang2024chatlogic,mensfelt2026prologmcp}. Most directly,
ASPBench evaluates LLMs on ASP entailment, answer-set verification, and answer-set
computation, finding models handle the first two but struggle to compute
answer sets \citep{ren2026aspbench}. \NAFBench differs in intent: rather than
measuring end-to-end ASP task-solving, we hold the program fixed and isolate
whether a model follows a specified negation semantics when several
legitimate ones diverge, and we use a translate-then-solve condition as a
mitigation baseline that cleanly separates reading a rulebook from applying
it.

\paragraph{Controlled natural language.}
Our verbalizations sit on the spectrum between fully natural language and formal
logic, ranging from abstract proposition-level conditionals to structured
narrative templates reminiscent of controlled natural languages and Logical
English \citep{kowalski2022logicalenglish}. This
lets us vary surface framing while preserving formal content, and connects the
benchmark to the practical interfaces (legal and regulatory rulebooks written in
controlled English).

\paragraph{Improving logical reasoning by data and training.}
A complementary line seeks to improve rather than merely measure logical
competence. Logic-driven data augmentation converts text into abstract meaning
representations and applies logically-equivalent transformations -- including
negation and contraposition -- to enlarge training sets
\citep{bao2024amrlda}; preference optimization can balance logical grounding
against fluency \citep{bao2026rlearner}; and structured cognitive priors can
mitigate the ``logic inertia'' by which models override stated rules with prior
beliefs \citep{bao2025conflict}. These training-side interventions motivate our own mitigation study: we use
\NAFBench{} not only to diagnose default correspondence but to test whether
translate-then-solve prompting and fine-tuning across framings restore
semantics-following.

\section{Discussion}

With the semantics left unspecified, two frontier models converge on a coherent
policy: they assert a value only when the negation cycle has at least one
consistent resolution and that value is shared by all resolutions, declining
otherwise. This policy matches none of the four certified viewpoints, but the
models agree on every unspecified item, and no constant-answer baseline
reproduces it. Once a viewpoint is named, both follow it correctly at the
evaluated program size. o4-mini instead drops under WFS, mainly by collapsing
undefined to false, connecting its errors to the broader difficulty of warranted
abstention. The open panel is substantially weaker. Fine-tuning improves
performance but does not transfer to an unseen abstract framing, while
autoformalization remains promising.

\section{Limitations}

Instances are synthetic, and the frontier models are evaluated on a fixed-complexity cell, so the results do not establish competence over arbitrary programs. We leave deeper programs, interacting cycles, other forms of negation, and real-world validation to future work. Trace analysis is approximate because it relies on pattern-based extraction.
\section{Conclusion and Future Work}

We introduced \NAFBench, a solver-certified generator for testing whether models apply a specified reading of default negation. Following the requested reading remains unsolved below the frontier, while our mitigations localize the gap to applying the rulebook. Future work will expand program complexity, improve trace analysis, and evaluate transfer to real-world legal, medical, and regulatory tasks.

\section*{Acknowledgments}
This work was supported by a Leverhulme Trust International Professorship Grant (LIP-2022-001).

\bibliography{references}

\clearpage
\appendix
\onecolumn
\section*{Technical Appendix}

\subsection{Public leaderboard and released dataset}
\label{sec:leaderboard}

\paragraph{Released dataset.}
We release \NAFBench{} as a public benchmark from the
solver-certified generator.
The scored task \tid{hard} is a
single set of $77$ programs / $385$ prompts.
A single
\tid{load\_dataset(``qbao775/naf-bench'')} returns three seed-disjoint splits:
\tid{train} ($960$, with gold), \tid{validation} ($385$ dev, with gold), and
\tid{test} ($385$, gold withheld); fresh seeds keep it regenerable and
contamination-free.

\paragraph{Instance construction.}
The released \tid{hard} set is generated differently from the production set used
in the main paper. That set is organised by \emph{divergence bin} (a
propagation-decidable control plus three negation-cycle families engineered so the
four semantics disagree); the leaderboard generator instead draws from six
solver-certified \emph{families}, chosen so that (i)~the gold answer cannot be
guessed from the prompt condition---each family contributes a different mix of
certified $(\mathrm{cred},\mathrm{skept})$ signatures---and (ii)~at least one
family poses a genuine search problem rather than a propagation-decidable one.
Every instance is a normal logic program whose atoms, stable models, and query
verdict are certified under all four semantics (SWI-Prolog, a well-founded solver,
and clingo). The families are:
\begin{itemize}\setlength{\itemsep}{2pt}
\item \tid{cnf} (search): a random 3-SAT instance near the satisfiability phase
transition ($m/n \approx 4.26$), embedded as $n$ independent choice cycles with one
integrity constraint per clause that discards every model violating that clause;
its stable models are exactly the satisfying assignments. The query asks about a
single variable, so its credulous truth is equivalent to the CNF being satisfiable
with that literal and its skeptical truth to the CNF being unsatisfiable with the
opposite literal---both requiring search. \tid{cnf\_n8} uses $8$ variables.
\item \tid{decided}: an acyclic, stratified program whose query is definitely true
or false. It is the only family that yields a definite \tid{A}/\tid{B} under the
closed-world and well-founded readings (the cyclic families all give \tid{C});
stack depths $1$--$6$.
\item \tid{loopy}: a negation cycle on which the well-founded reading decides the
query while SLDNF diverges, separating the two closed-world-style readings; loop
depths $1$--$4$.
\item \tid{parity}: the query holds iff an even number of $n$ independent cycles
selected a designated atom, accumulated through a stratified parity gadget, so no
prefix of the choices settles the answer; $n \in \{4,6,8\}$.
\item \tid{coupled}: interdependent constraints over shared atoms for which no
local assignment decides the query; $n \in \{3,5\}$.
\item \tid{control}: a propagation-decidable (low-ratio, satisfiable) instance
padded with gold-neutral filler to match the token length of a search tier, so that
any effort difference can be attributed to search rather than to reading a longer
prompt.
\end{itemize}

\paragraph{Public leaderboard.}
A free, automated competition accompanies it: a \tid{\{id, prediction\}} JSONL
(optionally with a \tid{trace}), opened as a pull request, is scored server-side
against the private \tid{test} gold and auto-merged when it touches only
submissions, so labels are never exposed. Ranking is by JOINT. Traced submissions
also get \emph{trace-sound}---the share of a model's \emph{correct} answers whose
trace commits to the certified query verdict---from the regex parser above, hence it is
an approximate, auxiliary signal, not a ranking criterion.

\paragraph{Current standings.}
Table~\ref{tab:leaderboard} gives our baselines on the hidden \tid{test} split.
The JOINT/trace-sound gap is the point: the explicit reasoners (Gemma4,
DeepSeek-R1) justify most correct answers, Qwen2.5-coder reasons soundly under
half the time, and the program-blind constant baseline has no trace to verify.

\medskip
{\centering
\setlength{\tabcolsep}{3pt}\footnotesize
\begin{tabular}{@{}lcccccc@{}}
\toprule
Model & JOINT & \shortstack{trace-\\sound$^{a}$} & sldnf & cred & skept & wfs \\
\midrule
Gemma4-31B         & \textbf{75.3} & 84.9  & 100.0 & 85.7 & 77.9 & 98.7 \\
DeepSeek-R1-32B    & 41.6          & 72.2  & 85.7  & 62.3 & 58.4 & 64.9 \\
Qwen2.5-coder-32B  & 31.2          & 41.8  & 74.0  & 59.7 & 59.7 & 67.5 \\
constant baseline  & 22.1          & ---   & 61.0  & 57.1 & 61.0 & 35.1 \\
Llama3-8B          & 2.6           & 24.6  & 35.1  & 51.9 & 42.9 & 44.2 \\
\bottomrule
\end{tabular}
\par}
\captionof{table}{Leaderboard standings on the hidden \tid{test} split ($385$ prompts,
$77$ programs); accuracy (\%), ranked by JOINT. $^{a}$\emph{trace-sound} is an
auxiliary regex-based approximation, not a ranking criterion.}
\label{tab:leaderboard}
\medskip

\subsection{Default Reading}

The default-reading analysis uses a prompt that specifies no semantics, asking
only for commonsense reasoning. Its 120 prompts therefore carry no gold answer
and serve instead to identify which reading an uninstructed model's answers
follow. Agreement with each candidate reading is computed for the models and,
alongside them, for constant-answer baselines.

\begin{table}[htb]
\centering
\small
\setlength{\tabcolsep}{2pt}
\newcommand{\bingap}{\hspace{12pt}}
\begin{tabular}{l @{\bingap} rrr @{\bingap} rrr @{\bingap} rrr @{\bingap} rrr}
\toprule
Model & \multicolumn{3}{c}{\tid{control}} & \multicolumn{3}{c}{\tid{ev\_one}}
      & \multicolumn{3}{c}{\tid{odd}} & \multicolumn{3}{c}{\tid{ev\_both}} \\
\midrule
Claude Sonnet 5   & 30A &     &     &     &     & 30C &     &     & 30C & 30A &    &    \\
GPT-5.6 Sol       & 30A &     &     &     &     & 30C &     &     & 30C & 30A &    &    \\
o4-mini           & 12A &     & 18C &     &  1B & 29C &  1A &  2B & 27C & 30A &    &    \\
\midrule
Qwen2.5-coder 32B & 10A &     & 20C & 19A &  2B &  9C & 15A &  2B & 12C & 22A &    & 8C \\
DeepSeek-R1 32B   &  7A & 1B  & 22C & 17A &  5B &  6C &  8A & 10B &  8C & 23A & 1B & 1C \\
Llama3-8B         & 19A &     & 11C & 27A &  1B &  2C & 21A &     &  9C & 22A & 1B & 7C \\
\midrule
\multicolumn{13}{l}{\emph{letter each candidate reading requires}} \\
credulous        & \multicolumn{3}{c}{A} & \multicolumn{3}{c}{A} & \multicolumn{3}{c}{B} & \multicolumn{3}{c}{A} \\
skeptical        & \multicolumn{3}{c}{A} & \multicolumn{3}{c}{B} & \multicolumn{3}{c}{A} & \multicolumn{3}{c}{A} \\
well-founded     & \multicolumn{3}{c}{A} & \multicolumn{3}{c}{C} & \multicolumn{3}{c}{C} & \multicolumn{3}{c}{C} \\
closed-world     & \multicolumn{3}{c}{A} & \multicolumn{3}{c}{C} & \multicolumn{3}{c}{C} & \multicolumn{3}{c}{C} \\
stable-consensus & \multicolumn{3}{c}{A} & \multicolumn{3}{c}{C} & \multicolumn{3}{c}{C} & \multicolumn{3}{c}{\textbf{A}} \\
\bottomrule
\end{tabular}
\caption{Answers under the unspecified-semantics condition (\tid{none}) per
divergence bin, with 30 programs in each bin. A $=$ definitely yes, B $=$
definitely no, and C $=$ cannot be determined; blank subcolumns are zero, and
row totals below 30 indicate unparsed answers.}
\label{tab:untutored}
\end{table}

Table~\ref{tab:untutored} exposes a striking difference between the two model
panels. Claude Sonnet 5 and GPT-5.6 Sol give exactly the same answer in every
bin: they answer definitively when the cycle has a consistent resolution shared
across the relevant alternatives, and return \textit{cannot be determined} when the alternatives disagree or
no stable resolution exists. This pattern is coherent, but it is not identical
to any of the four certified readings. o4-mini approximates the same pattern
with several deviations. The open models are more heterogeneous: their answers
vary substantially both across bins and within a bin, so no single qualitative
default describes the whole panel.

\begin{table}[htb]
\centering
\small
\setlength{\tabcolsep}{4pt}
\begin{tabular}{lcccc}
\toprule
Model & cred & skept & wfs & c-world \\
\midrule
Claude Sonnet 5   & 60 & 60 & 90 & 90 \\
GPT-5.6 Sol       & 60 & 60 & 90 & 90 \\
o4-mini           & 44 & 44 & 68 & 68 \\
\midrule
Qwen2.5-coder 32B & 53 & 49 & 39 & 39 \\
DeepSeek-R1 32B   & 57 & 43 & 22 & 22 \\
Llama3-8B         & 68 & 63 & 37 & 37 \\
\midrule
\emph{always A}   & 90 & 90 & 30 & 30 \\
\emph{always B}   & 30 & 30 & \phantom{0}0 & \phantom{0}0 \\
\emph{always C}   & \phantom{0}0 & \phantom{0}0 & 90 & 90 \\
\bottomrule
\end{tabular}
\caption{Number of answers, out of 120 prompts, agreeing with each candidate
reading under the unspecified-semantics condition. Unparsed answers agree with
no reading.}
\label{tab:fit}
\end{table}

Table~\ref{tab:fit} makes the same point quantitatively. Even the best-fitting
certified reading explains only part of each model's uninstructed behaviour.
For the two strongest frontier models, WFS and closed-world each agree on 90 of
120 prompts, yet Table~\ref{tab:untutored} shows that neither explains the
even-both-sided bin: both require C there, whereas the models uniformly answer
A. The constant baselines also fail to reproduce their bin-dependent pattern.
For the open models, agreement is lower and distributed across multiple
readings, supporting a behavioural description of their defaults rather than
attributing a single formal semantics to them.

\subsection{Trace Analysis}

The audit proceeds in four steps. First, each instance is rebuilt from its recorded seed and asserted equal to the stored program text, and the rebuilt program is solved under all four readings---the alternating fixpoint for well-founded, clingo stable models for credulous and skeptical entailment, and an analytic label for closed-world. Second, we audit
\tid{cq}, the atom at which the negation cycle meets the positive chain and the
only atom whose value separates the four readings, and $q$, the query to which
that value must propagate. The chain and purely positive block are skipped:
a given $t_i$'s value is instance-dependent, and no model fails the positive
block. Third, trace prose is normalised and split into clauses. Clauses making no
global claim are discarded (rule restatements, subgoal announcements,
hypotheses, and model-relative assertions), and the last surviving verdict on
each target atom is recorded together with three structural flags. Finally,
each trace is classified as sound, contradicted, unverifiable, or no derivation.
Only sound and contradicted traces enter the reported rate; unverifiable traces
and responses containing no derivation are excluded. The reported rates are
therefore conditional on the auditable subset.

\begin{table}[htbp]
\centering
\small
\setlength{\tabcolsep}{4pt}
\begin{tabular}{lrrrr}
\toprule
Model & correct & auditable & sound & contradicted \\
\midrule
Claude Sonnet 5   & 480 & 219 & 219 & \phantom{0}0 \ (0.0\%) \\
GPT-5.6 Sol       & 480 & 262 & 262 & \phantom{0}0 \ (0.0\%) \\
o4-mini           & 455 & 243 & 243 & \phantom{0}0 \ (0.0\%) \\
\midrule
Qwen2.5-coder 32B & 314 & 150 & 129 & 21 \ (14.0\%) \\
DeepSeek-R1 32B   & 224 & 105 & \phantom{0}81 & 24 \ (22.9\%) \\
Llama3-8B         & 224 & \phantom{0}71 & \phantom{0}43 & 28 \ (39.4\%) \\
\bottomrule
\end{tabular}
\caption{Among correct answers, the frequency with which an auditable trace
states a verdict that contradicts solver certification.}
\label{tab:audit-correct}
\end{table}

Answer accuracy alone can overstate semantic competence because a model may
arrive at the correct option through a derivation that contradicts the
certified intermediate values. Table~\ref{tab:audit-correct} measures this
``right answer, wrong derivation'' failure among correct answers whose traces
make an auditable claim. No such contradiction is found for the three frontier
models in this subset. In contrast, the contradiction rate rises from 14.0\%
for Qwen2.5-coder to 22.9\% for DeepSeek-R1 and 39.4\% for Llama3. The remaining
correct traces are not automatically counted as sound: traces that state no
checkable verdict are excluded rather than being treated as evidence.

\begin{table}[htbp]
\centering
\small
\setlength{\tabcolsep}{3pt}
\begin{tabular}{lcccc}
\toprule
Model & \tid{sldnf} & \tid{cred} & \tid{skept} & \tid{wfs} \\
\midrule
Claude Sonnet 5   & 42/42 & 48/48 & 59/59 & 70/70 \\
GPT-5.6 Sol       & 27/27 & 42/42 & 95/95 & 98/98 \\
o4-mini           & 42/42 & 67/67 & 61/61 & \textbf{74/96} \\
\midrule
Qwen2.5-coder 32B & 17/54 & 37/55 & 34/76 & 46/67 \\
DeepSeek-R1 32B   & 26/95 & 12/25 & 23/42 & 24/94 \\
Llama3-8B         & \phantom{0}6/43 & 10/33 & 17/59 & 15/62 \\
\bottomrule
\end{tabular}
\caption{Sound derivations over auditable traces, by prompting condition.}
\label{tab:audit-cond}
\end{table}

The condition-level breakdown in Table~\ref{tab:audit-cond} localises these
failures. Claude Sonnet 5 and GPT-5.6 Sol are sound on every auditable trace in
all four conditions. o4-mini is likewise perfect under SLDNF and both
stable-model readings, but falls to $74/96$ under WFS, consistent with its
tendency to collapse an undefined value to false. The open models show a wider
gap between producing an answer and correctly applying the requested
rulebook; their sound-derivation rates vary markedly across conditions.

\begin{table}[htbp]
\centering
\small
\setlength{\tabcolsep}{5pt}
\begin{tabular}{lcc}
\toprule
& \tid{odd}$+$\tid{skept} & \tid{odd}$+$\tid{cred} \\
Model & correct / earned & correct / earned \\
\midrule
Claude Sonnet 5   & 30 / 30 & 30 / 30 \\
GPT-5.6 Sol       & 30 / 30 & 30 / 30 \\
o4-mini           & 30 / 29 & 30 / 30 \\
\midrule
Qwen2.5-coder 32B & 23 / \textbf{\phantom{0}1} & \phantom{0}3 / 1 \\
DeepSeek-R1 32B   & 14 / 10 & 18 / 5 \\
Llama3-8B         & 25 / \textbf{\phantom{0}0} & \phantom{0}2 / 1 \\
\bottomrule
\end{tabular}
\caption{``Earned'' means that the trace states that the program has no stable
model---the reason the skeptical answer is vacuously yes. Earned status is detected by pattern-matching the trace for an explicit no-stable-model statement, so an unmatched phrasing reads as unearned; the earned counts are lower bounds.}
\label{tab:odd}
\end{table}

The odd-cycle case illustrates why this distinction matters. An odd negative
cycle has no stable model, so credulous entailment is false while skeptical
entailment is vacuously true. A model that overlooks the absence of stable
models can nevertheless obtain the skeptical answer by ordinary forward
chaining. Table~\ref{tab:odd} therefore distinguishes a correct letter from an
\emph{earned} answer whose trace states the decisive fact. Nearly all frontier
answers are earned, whereas only 1 of Qwen's 23 correct skeptical answers and
none of Llama3's 25 are supported by the required no-stable-model reasoning.
The corresponding credulous column removes this accidental advantage and
shows the sharp reduction in correct answers for those models.


\subsection{Expanded Mitigation Analysis}

The main paper reports mitigation results on a structurally diverse set of 44
WFS programs, replacing the earlier 12-item diagnostic probe. The expanded set
covers even negation cycles with $k\in\{2,4,6\}$, conjunctive and disjunctive
query support, and control programs, with multiple structurally distinct
instances for each parameter setting. Table~\ref{tab:mit44} provides the full
per-model results for the three open models under three settings: \emph{direct}
(the baseline prompt), \emph{verify} (a prompt-only verify-before-answer
scaffold), and \emph{T2S} (translate-then-solve, with the translated program
scored by the certified solver).

We report both parsed accuracy, computed over responses from which a final
answer can be extracted, and strict accuracy, which counts unparseable or
non-committal responses as incorrect. This distinction is especially important
for DeepSeek-R1, which produces substantially more null outputs than the other
models.

\begin{table}[htbp]
\centering
\small
\setlength{\tabcolsep}{5pt}
\begin{tabular}{llccc}
\toprule
Model & Metric & Direct & Verify & T2S \\
\midrule
\multirow{2}{*}{Qwen2.5-coder 32B}
 & Parsed accuracy (\%) & 59 & 82 & 95 \\
 & Strict accuracy (\%; nulls) & 59 (0) & 82 (0) & 95 (0) \\
\midrule
\multirow{2}{*}{DeepSeek-R1 32B}
 & Parsed accuracy (\%) & 62 & 69 & 89 \\
 & Strict accuracy (\%; nulls) & 45 (12) & 57 (8) & 77 (6) \\
\midrule
\multirow{2}{*}{Llama3-8B}
 & Parsed accuracy (\%) & 49 & 62 & 74 \\
 & Strict accuracy (\%; nulls) & 48 (1) & 48 (10) & 70 (2) \\
\bottomrule
\end{tabular}
\caption{Mitigation results on the 44-program WFS set. Null counts are shown in
parentheses for strict accuracy. Translation success for T2S was $44/44$ for
Qwen2.5-coder, $38/44$ for DeepSeek-R1, and $42/44$ for Llama3; unsuccessful
translations are counted as incorrect under strict scoring.}
\label{tab:mit44}
\end{table}

Both mitigation strategies improve parsed accuracy over direct prompting for
every open model. The verify scaffold yields gains of 7--23 percentage points,
showing that an explicit intermediate check can reduce some semantic errors
without external tools. T2S is consistently stronger, improving parsed accuracy
by 25--36 points and reaching 95\% for Qwen2.5-coder. Its remaining errors arise
from both autoformalization failures and incorrect or unparseable translations;
when translation succeeds, the certified solver removes the downstream burden
of applying WFS correctly. The parsed--strict gap also reveals that apparent
reasoning quality can be overstated when abstentions or non-committal outputs are
excluded, most notably for DeepSeek-R1.

\subsubsection{Mitigation Prompt Templates}

For reproducibility, we provide the two prompt interventions verbatim below.
The numbering follows the mitigation numbering used in the main paper.
Mitigation~1 asks the model only to autoformalize the natural-language scenario,
after which the certified solver applies WFS. Mitigation~3 retains end-to-end
answer generation but requires the model to state the semantics, assign a truth
value to every atom, and only then evaluate the query. Both prompts were applied
to the same 44-program set used in Table~\ref{tab:mit44}.

\smallskip
\noindent\textbf{Mitigation 1 --- translate-then-solve (system prompt).}
\begin{Verbatim}[breaklines=true,breakanywhere=true,fontsize=\small]
You translate natural-language rule sets into a
 ground normal logic program. Output ONLY the
 program, one statement per line, in this grammar:
   fact.
   head :- b1, b2, not c1.
   QUERY: <the queried atom>
Use short ground atom names (no variables). Encode
'X holds if and only if Y does not' as the rule
'x :- not y.' (one rule per such statement). Encode
'P unless Q' as 'p :- <conditions>, not q.'. Do not
solve or explain; output only the program and the
QUERY line.
\end{Verbatim}

\smallskip
\noindent\textbf{Mitigation 3 --- verify-before-answer scaffold.}
\begin{Verbatim}[breaklines=true,breakanywhere=true,fontsize=\small]
Before answering, reason in three explicit,
labelled steps:
STEP 1 - State the exact semantics you must apply
and its rule for 'true'/'false'/'undefined'.
STEP 2 - For EACH atom that appears, determine its
truth value under THAT semantics. For any atom whose
support runs through a cycle of negations, check
whether it is grounded in facts; if it is not, it is
'undefined' (do NOT case-split into separate
consistent worlds unless the semantics tells you to).
STEP 3 - Evaluate the queried atom from the Step-2
values (undefined propagates: anything that depends
on an undefined atom and is not otherwise grounded is
itself undefined).
Then give your final answer as a line 'ANSWER: X'
where X is A, B, or C.
\end{Verbatim}


\subsection{Example Traces}

The following examples make the audit categories concrete. For each model, a
\emph{sound} example states intermediate and final verdicts consistent with the
certified reading. A \emph{contradicted} example contains at least one explicit
claim about \tid{cq} or $q$ that conflicts with certification, irrespective of
whether its final answer letter happens to be correct. An \emph{untutored}
example comes from the unspecified-semantics condition and illustrates the
model's default reasoning behaviour. We include the stored traces verbatim so
that the classifications and the limits of the automated extractor can be
inspected directly.

\subsubsection{Claude Sonnet 5}

\paragraph{Sound}
\input{trace-examples/claude-sonnet-5-sound}

\paragraph{Untutored}
\input{trace-examples/claude-sonnet-5-none}

\paragraph{GPT-5.6 Sol}

\subsubsection{Sound}
\input{trace-examples/gpt-5-6-sol-sound}

\paragraph{Untutored}
\input{trace-examples/gpt-5-6-sol-none}

\subsubsection{o4-mini}

\paragraph{Sound}
\input{trace-examples/o4-mini-sound}

\paragraph{Contradicted}
\input{trace-examples/o4-mini-unsound}

\paragraph{Untutored}
\input{trace-examples/o4-mini-none}

\subsubsection{Qwen2.5-coder 32B}

\paragraph{Sound}
\input{trace-examples/qwen2-5-coder-32b-sound}

\paragraph{Contradicted}
\input{trace-examples/qwen2-5-coder-32b-unsound}

\paragraph{Untutored}
\input{trace-examples/qwen2-5-coder-32b-none}

\subsubsection{DeepSeek-R1 32B}

\paragraph{Sound}
\input{trace-examples/deepseek-r1-32b-sound}

\paragraph{Contradicted}
\input{trace-examples/deepseek-r1-32b-unsound}

\paragraph{Untutored}
\input{trace-examples/deepseek-r1-32b-none}

\subsubsection{Llama3-8B}

\paragraph{Sound, and Why That Label Is Narrow}
\input{trace-examples/llama3-8b-sound}

\paragraph{Contradicted}
\input{trace-examples/llama3-8b-unsound}

\paragraph{Untutored}
\input{trace-examples/llama3-8b-none}

\end{document}

%% file: trace-examples/claude-sonnet-5-sound.tex
\begin{Verbatim}[breaklines=true,breakanywhere=true,fontsize=\scriptsize,frame=single,framesep=3pt,samepage=false]
Looking at this problem, I need to trace the dependency chain for q.

**Step 1: Trace q's dependency chain**
- q <- t0 <- t1 <- t2 <- t3 <- t4 <- (t5 and wide)
- t5 <- (t6 and cq)
- t6 <- t7 <- btrue (btrue is a fact, so t7=true, t6=true)

**Step 2: Evaluate cq**
- cq <- x0
- x0 <- not x1
- x1 <- not x2
- x2 <- not x0

This forms a cycle: x0 depends on not-x1, x1 depends on not-x2, x2 depends on not-x0 --- an odd-length cycle through negation. No stable assignment of true/false satisfies this consistently (odd negative cycle), so under well-founded semantics, x0, x1, x2 are all **undefined**.

Since cq's only rule depends positively on x0 (undefined), cq is also **undefined**.

**Step 3: Evaluate wide (independent chain)**
- s0 <- g0_0,g0_1,g0_2,g0_3 (all facts, true) -> s0 = true
- s1 <- g1_0,g1_1 (facts, true) -> s1 = true
- s2 <- g2_0 (fact, true) -> s2 = true
- s3 <- g3_0 (fact, true) -> s3 = true
- p0 <- s0,s1,s2 (all true) -> p0 = true
- p3 <- s2 (true) -> p3 = true
- p1 <- s0,s1,s3 (all true) -> p1 = true
- p2 <- s1,s2,s3 (all true) -> p2 = true
- wide <- p0,p1,p2,p3 (all true) -> **wide = true**

**Step 4: Combine back up the chain**
- t5 <- t6(true) and cq(undefined) -> t5 = **undefined**
- t4 <- t5(undefined) and wide(true) -> t4 = **undefined**
- t3 <- t4(undefined) -> t3 = undefined
- t2 <- t3(undefined) -> t2 = undefined
- t1 <- t2(undefined) -> t1 = undefined
- t0 <- t1(undefined) -> t0 = undefined
- q <- t0(undefined) -> **q = undefined**

Since q's truth value depends on an unresolved odd cycle through negation (via x0/x1/x2), q is undefined under well-founded semantics.

ANSWER: C
\end{Verbatim}

%% file: trace-examples/claude-sonnet-5-none.tex
\begin{Verbatim}[breaklines=true,breakanywhere=true,fontsize=\scriptsize,frame=single,framesep=3pt,samepage=false]
The chain from btrue leads through t7,t6,t5,t4,t3,t2 all being true (all trivially propagate since btrue is true).

For wide: s0 (g0_0/\g0_1), s1(g1_0/\g1_1/\g1_2), s2(g2_0), s3(g3_0/\g3_1) are all true given facts, so p0 and p1 are true, making wide true.

For cq: it depends on x0,x1,x2,x3, which form a self-referential loop based on etrue (true) and negations of each other. Testing consistent assignments:

- If x3=F: x2=T, x1=F, x0=T -> consistent, and x0=T,x2=T means cq is true (via cq-from-x0 or cq-from-x2 rules).
- If x3=T: x2=F, x1=T, x0=F -> also consistent, and x1=T,x3=T means cq is true (via cq-from-x1 or cq-from-x3 rules).

In both consistent fixed points, cq ends up true, so cq is true regardless of which valid assignment holds.

Since t2, wide, and cq are all true, t1 is true, hence t0 is true, hence q is true.

ANSWER: A
\end{Verbatim}

%% file: trace-examples/gpt-5-6-sol-sound.tex
\begin{Verbatim}[breaklines=true,breakanywhere=true,fontsize=\scriptsize,frame=single,framesep=3pt,samepage=false]
The cycle among \(x0, x1, x2\) through default negation leaves them undefined. Therefore \(cq\), and consequently the chain leading through \(t6\) to \(q\), is also undefined.

ANSWER: C
\end{Verbatim}

%% file: trace-examples/gpt-5-6-sol-none.tex
\begin{Verbatim}[breaklines=true,breakanywhere=true,fontsize=\scriptsize,frame=single,framesep=3pt,samepage=false]
The given facts derive \(s0,s1,s2,s3\), which derive all \(p0,p1,p2,p3\), and hence `wide`.

Since `btrue` is true, the chain derives \(t7,t6,t5,t4\). As `x3` is not true, `x2` is true, so `cq` is true. Therefore \(t3,t2,t1,t0\) and finally `q` are true.

ANSWER: A
\end{Verbatim}

%% file: trace-examples/o4-mini-sound.tex
\begin{Verbatim}[breaklines=true,breakanywhere=true,fontsize=\scriptsize,frame=single,framesep=3pt,samepage=false]
After grounding and computing answer sets, the only choice-points lie in the 4-cycle of x0,x1,x2,x3.  One stable model makes x0,x2 true, which in turn makes cq, t1, t0 and hence q true.  Thus q holds in at least one answer set.  
ANSWER: A
\end{Verbatim}

%% file: trace-examples/o4-mini-unsound.tex
\begin{Verbatim}[breaklines=true,breakanywhere=true,fontsize=\scriptsize,frame=single,framesep=3pt,samepage=false]
Let's evaluate in the well-founded model step by step:

1.  etrue, btrue, g*_ facts are all true.
2.  x0, x1, x2 form a default-negation cycle (x0<-etrue /\ not x1; x1<-etrue /\ not x2; x2<-etrue /\ not x0).  Under WFS they are all undefined.
3.  Hence cq<-x0 is undefined (no founded support), so cq is false (its only rule is inapplicable).
4.  t7<-btrue /\ cq fails (cq is false), so t7 is false.
5.  Then t6<-t7, t5<-t6, t4<-t5 are all false in turn.
6.  t3<-t4 /\ wide fails (t4 is false), so t3 is false; hence t2<-t3 and t1<-t2 are false.
7.  Finally t0<-t1 fails, so t0 and thus q<-t0 is false.

Since q is founded false, the answer is "Definitely no."

ANSWER: B
\end{Verbatim}

%% file: trace-examples/o4-mini-none.tex
\begin{Verbatim}[breaklines=true,breakanywhere=true,fontsize=\scriptsize,frame=single,framesep=3pt,samepage=false]
Think of x0,x1,x2,x3. Since etrue is true we have the 4-cycle
  x0 = ~x1,
  x1 = ~x2,
  x2 = ~x3,
  x3 = ~x0.
This forces two consistent assignments:
  Case1: x0=T,x1=F,x2=T,x3=F
  Case2: x0=F,x1=T,x2=F,x3=T.
In either case at least one of x0,x1,x2,x3 is true, so cq is true.  Then t4 (since t5, via t6,t7,w/ wide true by s0--s3 and p0--p3) is true, so t3,t2,t1,t0 all follow, yielding q true.  
ANSWER: A
\end{Verbatim}

%% file: trace-examples/qwen2-5-coder-32b-sound.tex
\begin{Verbatim}[breaklines=true,breakanywhere=true,fontsize=\scriptsize,frame=single,framesep=3pt,samepage=false]
To determine if proposition q is true, we need to trace back the dependencies of its truth value according to the given rules.

1. Proposition q is true if proposition t0 is true.
2. Proposition t0 is true if proposition t1 is true and proposition cq is true.
3. Proposition cq is true if proposition x0 is true.
4. Proposition x0 is true if proposition etrue is true and proposition x1 is not true.
5. Proposition etrue is true (given).
6. Proposition x1 is true if proposition etrue is true and proposition x2 is not true.
7. Proposition x2 is true if proposition etrue is true and proposition x0 is not true.

Let's analyze the truth values of x0, x1, and x2:
- Since etrue is true, the truth value of x0 depends on whether x1 is not true.
- The truth value of x1 depends on whether x2 is not true.
- The truth value of x2 depends on whether x0 is not true.

This creates a cycle: x0 -> x1 -> x2 -> x0. Since there's no external information to break this cycle, the well-founded semantics would classify x0, x1, and x2 as undefined because their truth values depend on each other in an unresolved manner.

Since x0 is undefined, cq (which depends on x0) is also undefined.
Since cq is undefined, t0 (which depends on cq) is also undefined.
Since t0 is undefined, q (which depends on t0) is also undefined.

Therefore, the truth value of proposition q cannot be determined based on the given information and rules.

ANSWER: C
\end{Verbatim}

%% file: trace-examples/qwen2-5-coder-32b-unsound.tex
\begin{Verbatim}[breaklines=true,breakanywhere=true,fontsize=\scriptsize,frame=single,framesep=3pt,samepage=false]
To determine if proposition q is true in every answer set, we need to trace the dependencies and conditions leading to its truth.

1. Proposition q is true if proposition t0 is true.
2. Proposition t0 is true if proposition t1 is true.
3. Proposition t1 is true if proposition t2 is true.
4. Proposition t2 is true if proposition t3 is true.
5. Proposition t3 is true if proposition t4 is true.
6. Proposition t4 is true if proposition t5 is true and proposition wide is true.
7. Proposition t5 is true if proposition t6 is true and proposition cq is true.
8. Proposition t6 is true if proposition t7 is true.
9. Proposition t7 is true if proposition btrue is true.
10. Proposition btrue is true (given).
11. Proposition wide is true if proposition p0, p1, p2, and p3 are all true.

Let's check the conditions for p0, p1, p2, and p3:
- Proposition p0 is true if s0, s1, and s2 are all true.
  - s0 is true if g0_0, g0_1, g0_2, and g0_3 are all true (all given).
  - s1 is true if g1_0 and g1_1 are both true (both given).
  - s2 is true if g2_0 is true (given).
- Proposition p1 is true if s0, s1, and s3 are all true.
  - We already know s0 and s1 are true.
  - s3 is true if g3_0 is true (given).
- Proposition p2 is true if s1, s2, and s3 are all true.
  - We already know s1, s2, and s3 are true.
- Proposition p3 is true if s2 is true.
  - s2 is true (given).

Since btrue is given as true, t7 is true. Since t7 is true, t6 is true. Since t6 and cq are both true, t5 is true. To determine cq, we need to check x0:
- Proposition x0 is true if x1 is not true.
- Proposition x1 is true if x2 is not true.
- Proposition x2 is true if x3 is not true.
- Proposition x3 is true if x0 is not true.

This creates a circular dependency, but we can determine that either x0 and x3 are both false or both true. However, for cq to be true (which is necessary for t5), x0 must be true (since x1 would then be false, making x2 true, which makes x3 false, consistent with x0 being true).

Since all conditions for p0, p1, p2, and p3 are met, wide is true. Since t4, t5, and t6 are all true, t3 is true. Since t3 is true, t2 is true. Since t2 is true, t1 is true. Since t1 is true, t0 is true. Since t0 is true, q is true.

Since we have shown that q must be true in the given conditions and there are no contradictions leading to a scenario where q would not be true, the answer is:

ANSWER: A
\end{Verbatim}

%% file: trace-examples/qwen2-5-coder-32b-none.tex
\begin{Verbatim}[breaklines=true,breakanywhere=true,fontsize=\scriptsize,frame=single,framesep=3pt,samepage=false]
To determine if proposition q is true, we need to trace back the conditions that lead to its truth according to the given rules.

1. Proposition q is true if proposition t0 is true.
2. Proposition t0 is true if proposition t1 is true.
3. Proposition t1 is true if proposition t2 is true.
4. Proposition t2 is true if proposition t3 is true.
5. Proposition t3 is true if proposition t4 is true.
6. Proposition t4 is true if proposition t5 is true.
7. Proposition t5 is true if proposition t6 is true.
8. Proposition t6 is true if proposition t7 is true and proposition wide is true.
9. Proposition t7 is true if proposition btrue is true and proposition cq is true.

Given that proposition btrue is true, we need to determine the truth of propositions cq and t7 for t6 to be true.

10. Proposition cq can be true under several conditions:
    - If proposition x1 is true.
    - If proposition x3 is true.
    - If proposition x2 is true.
    - If proposition x0 is true.

However, we need to check the consistency of these propositions with other given rules:

11. Proposition x2 is true if proposition x3 is not true.
12. Proposition x3 is true if proposition x0 is not true.
13. Proposition x0 is true if proposition x1 is not true.
14. Proposition x1 is true if proposition x2 is not true.

From these, we can deduce that:
- If x2 is true, then x3 must be false (from rule 11).
- If x3 is false, then x0 could be either true or false (from rule 12).
- If x0 is true, then x1 must be false (from rule 13).
- If x1 is false, then x2 can be true (from rule 14).

This creates a consistent scenario where x2 and x1 can both be true under certain conditions, but we need to check if cq being true leads to t7 being true.

15. Since btrue is given as true, for t7 to be true, cq must also be true.
16. We have established that cq can indeed be true under multiple scenarios (x0, x1, x2, or x3).

Given that we have a consistent scenario where cq can be true and btrue is true, t7 can be true, which makes t6 true, leading to t5 being true, then t4, t3, t2, t1, and finally t0 being true.

Since proposition q is true if proposition t0 is true, and we have shown that t0 can indeed be true under the given conditions, we conclude:

ANSWER: A
\end{Verbatim}

%% file: trace-examples/deepseek-r1-32b-sound.tex
\begin{Verbatim}[breaklines=true,breakanywhere=true,fontsize=\scriptsize,frame=single,framesep=3pt,samepage=false]
<think>
Okay, so I need to figure out if proposition q is true based on the given rules using Well-Founded semantics. Let me start by understanding all the propositions and their dependencies.

First, let's list out all the propositions mentioned:

- g0_3: True.
- x3: True if x0 is not true.
- x0: True if x1 is not true.
- cq: True if x0 is true.
- t7: True if btrue is true.
- btrue: True.
- t4: True if t5 is true.
- wide: True if p0, p1, and p2 are all true.
- g1_0: True.
- x2: True if x3 is not true.
- p2: True if s1 and s2 are both true.
- q: True if t0 is true.
- s1: True if g1_0 is true.
- g3_1: True.
- g0_1: True.
- g0_0: True.
- s2: True if g2_0 is true.
- t6: True if t7 and cq are both true.
- t5: True if t6 is true.
- p0: True if s0, s1, and s3 are all true.
- t3: True if t4 is true.
- g3_0: True.
- t2: True if t3 is true.
- p1: True if s0, s1, s2, and s3 are all true.
- x1: True if x2 is not true.
- t1: True if t2 is true.
- g0_2: True.
- s0: True if g0_0, g0_1, g0_2, and g0_3 are all true.
- t0: True if t1 and wide are both true.
- s3: True if g3_0 and g3_1 are both true.
- g2_0: True.

Alright, that's a lot. Let me try to break it down step by step.

Starting with the propositions that are directly given as true:

- g0_3 is true.
- btrue is true.
- g1_0 is true.
- g3_1 is true.
- g0_1 is true.
- g0_0 is true.
- g2_0 is true.
- g3_0 is true.
- g0_2 is true.

Now, let's see what other propositions can be determined based on these.

First, s0: It depends on g0_0, g0_1, g0_2, and g0_3. All of these are true, so s0 is true.

s1 depends on g1_0, which is true, so s1 is true.

s3 depends on g3_0 and g3_1, both true, so s3 is true.

s2 depends on g2_0, which is true, so s2 is true.

Now, let's look at p0: It requires s0, s1, and s3. All are true, so p0 is true.

p1 requires s0, s1, s2, and s3. All are true, so p1 is true.

p2 requires s1 and s2. Both are true, so p2 is true.

Now, wide depends on p0, p1, and p2. All are true, so wide is true.

Next, let's look at the x propositions:

x3: True if x0 is not true. But we don't know x0 yet.

x0: True if x1 is not true. Again, depends on x1.

x1: True if x2 is not true.

x2: True if x3 is not true.

So this forms a cycle:

x3 <-> ~x0

x0 <-> ~x1

x1 <-> ~x2

x2 <-> ~x3

This creates an undefined situation because each depends on the negation of another, leading to a loop without any grounded truth. So x3, x0, x1, and x2 are all undefined.

Since x0 is undefined, cq (which requires x0) is also undefined.

Now, looking at t7: It's true if btrue is true, which it is. So t7 is true.

t6 depends on t7 and cq. Since t7 is true but cq is undefined, t6 is undefined because one of its conditions is undefined.

t5 depends on t6. Since t6 is undefined, t5 is also undefined.

t4 depends on t5. Since t5 is undefined, t4 is undefined.

t3 depends on t4. So t3 is undefined.

t2 depends on t3. So t2 is undefined.

t1 depends on t2. So t1 is undefined.

Now, t0 requires t1 and wide. t1 is undefined, but wide is true. However, since one condition (t1) is undefined, the entire proposition t0 becomes undefined because it's not fully supported.

Finally, q depends on t0. Since t0 is undefined, q is also undefined.

Wait a minute, let me double-check that. Because in Well-Founded semantics, if any part of a rule is undefined, does that make the whole rule undefined? Or can some parts be true even if others are undefined?

In this case, for t0 to be true, both t1 and wide must be true. Since t1 is undefined, we don't have enough information to conclude t0 as true or false. So t0 is undefined.

Therefore, q depends on t0 being true. But since t0 is undefined, q cannot be determined as either true or false. It remains undefined.
</think>

The evaluation of the propositions leads to a cycle in the x variables, making them undefined. Consequently, cq and several t variables are also undefined. Since t0 requires an undefined variable (t1), it becomes undefined, leaving q undefined.

**Answer:** The value of q is undefined.
\end{Verbatim}

%% file: trace-examples/deepseek-r1-32b-unsound.tex
\begin{Verbatim}[breaklines=true,breakanywhere=true,fontsize=\scriptsize,frame=single,framesep=3pt,samepage=false]
<think>
Okay, so I need to figure out if proposition q is true based on the given rules. Let me start by understanding what each rule says and how they connect.

First, looking at the question: Is q true? The rules say that q is true if t0 is true. So, my first step is to check whether t0 is true.

t0 is true if t1 is true. So now I need to see if t1 is true. Looking at the rules, t1 is true if t2 is true. Then, t2 is true if t3 is true. Next, t3 is true if t4 is true. Moving on, t4 is true if t5 is true.

Now, what makes t5 true? The rule says t5 is true if t6 is true, wide is true, and cq is true. So I need to check all three: t6, wide, and cq.

Starting with t6: It's true if t7 is true. And t7 is true if btrue is true. Since the rules state that btrue is true, t7 must be true, which makes t6 true.

Next, checking wide: It's true if p0, p1, p2, and p3 are all true. Let me look at each of these.

p0 is true if s1 and s2 are true. p1 is true if s1 and s3 are true. p2 requires s0, s1, s2, and s3 to be true. p3 needs s0, s1, and s2 to be true.

So I need to check each of these s propositions.

s0 is true if g0_0 is true, which it is because the rules say g0_0 is true.

s1 is true if g1_0, g1_1, and g1_2 are all true. The rules state that g1_0, g1_1, and g1_2 are all true, so s1 is true.

s2 is true if g2_0 is true, which it is.

s3 is true if g3_0, g3_1, and g3_2 are all true. The rules say these are all true, so s3 is true.

Now, let's check each p:

p0: s1 (true) and s2 (true) -> p0 is true.
p1: s1 (true) and s3 (true) -> p1 is true.
p2: s0 (true), s1 (true), s2 (true), s3 (true) -> p2 is true.
p3: s0 (true), s1 (true), s2 (true) -> p3 is true.

Since all four p's are true, wide is true.

Now, checking cq. The rules say that cq is true if x0 is true, or x3 is true, or x1 is true, or x2 is true. Let me see which of these apply.

Looking at the definitions:

x0 is true if x1 is not true.
x1 is true if x2 is not true.
x2 is true if x3 is not true.
x3 is true if x0 is not true.

This seems like a circular dependency. Let's try to evaluate each step by step.

Start with x0: It depends on whether x1 is false. But x1 is true only if x2 is false. So, let me see:

Assume x0 is true -> then x1 must be false.
If x1 is false, then according to its rule (x1 is true iff x2 is not true), since x1 is false, it means that x2 is true.

Now, if x2 is true, then according to x2's rule (x2 is true iff x3 is not true), x3 must be false.

If x3 is false, then according to x3's rule (x3 is true iff x0 is not true), since x3 is false, it means that x0 is true. Which brings us back to the initial assumption that x0 is true.

So this creates a loop where x0 is true -> x1 is false -> x2 is true -> x3 is false -> x0 is true. So in this case, x0 is true, which makes cq true because one of its conditions (x0) is met.

Alternatively, let's check if any other condition for cq is true:

- If x3 is true: But from above, x3 is false.
- If x1 is true: From above, x1 is false.
- If x2 is true: Yes, because x2 is true as per the earlier step.

Wait, but in the rules, cq can be true if any of these are true. So even if x0 is true, that's enough for cq to be true. But let me make sure I'm not missing anything else.

So, since x0 is true (as we've established), cq is definitely true because one of its conditions is satisfied.

Now, going back to t5: It requires t6 (true), wide (true), and cq (true). All three are true, so t5 is true. Then t4 is true because it depends on t5. Similarly, t3 is true because t4 is true, t2 is true because t3 is true, t1 is true because t2 is true, and finally t0 is true because t1 is true.

Since q is true if t0 is true, and we've established that t0 is true, then q must be true.
</think>

The evaluation shows that all necessary conditions for q being true are satisfied. 

ANSWER: A
\end{Verbatim}

%% file: trace-examples/deepseek-r1-32b-none.tex
\begin{Verbatim}[breaklines=true,breakanywhere=true,fontsize=\scriptsize,frame=single,framesep=3pt,samepage=false]
<think>
Okay, so I need to figure out if proposition q is true based on the given rules. Let me start by understanding each rule and see how they connect.

First, let's list all the propositions that are directly stated as true:

- g0_0 is true.
- g2_0 is true.
- g1_0 is true.
- g3_0 is true.
- g1_1 is true.
- g1_4 is true.
- g1_2 is true.
- btrue is true.

Now, let's look at the other propositions and see how they depend on these or others.

Starting with q: The rule says q is true if t0 is true. So I need to check if t0 is true.

t0 is true if t1 is true. So now I need to find out if t1 is true.

t1 is true if t2 is true. Next, t2 is true if t3 is true.

t3 is true if t4 is true. Then, t4 is true if t5 is true.

t5 is true if t6 is true. So now I need to check t6.

t6 is true if both t7 and wide are true. Let's look at t7 first.

t7 is true if btrue is true and cq is true. We know btrue is true, so we just need to see if cq is true.

Looking at cq: It has multiple conditions where it can be true. The rules say:

- cq is true if x1 is true.
- cq is true if x3 is true.
- cq is true if x0 is true.
- cq is true if x2 is true.

So, any of these being true would make cq true. Let's check each one.

First, x0: The rule says x0 is true if x1 is not true. So x0 depends on x1.

x1 is true if x2 is not true. So x1 depends on x2.

x2 is true if x3 is not true. So x2 depends on x3.

x3 is true if x0 is not true. Hmm, this creates a loop: x3 depends on x0, which depends on x1, which depends on x2, which depends on x3. Let's try to unravel this.

Let me assume that x3 is false. Then, according to the rule for x2, if x3 is not true (which it isn't), then x2 is true. So x2 would be true.

If x2 is true, then looking at x1: x1 is true if x2 is not true. But x2 is true, so x1 is false.

x0 is true if x1 is not true. Since x1 is false, x0 is true.

Now, let's check x3 again: x3 is true if x0 is not true. But we just found that x0 is true, so x3 must be false. That matches our initial assumption. So this loop holds.

So the values are:

- x3 = false
- x2 = true (because x3 is false)
- x1 = false (because x2 is true)
- x0 = true (because x1 is false)

Now, let's check cq again. Since x2 is true, and one of the conditions for cq is that if x2 is true, then cq is true. So cq is definitely true.

So t7 requires btrue (which is true) and cq (which we just found to be true). Therefore, t7 is true.

Now, going back to t6: it needs both t7 and wide to be true. We know t7 is true, so now check wide.

wide is true if p0 and p1 are both true. Let's look at p0 and p1.

p0 is true if s0, s1, s2, and s3 are all true.

s0: The rule says s0 is true if g0_0 is true. Since g0_0 is true, s0 is true.

s1: It's true if g1_0, g1_1, g1_2, g1_3, and g1_4 are all true. We know all these g1_* propositions are true (g1_0, g1_1, g1_2, g1_3, g1_4), so s1 is true.

s2: It's true if g2_0 is true. Since g2_0 is true, s2 is true.

s3: It's true if g3_0 is true. Since g3_0 is true, s3 is true.

So p0 requires all four s propositions to be true, which they are. Therefore, p0 is true.

Now check p1: It's also true if s0, s1, s2, and s3 are all true. We've already established that each of these is true, so p1 is true as well.

Since both p0 and p1 are true, wide is true.

So t6 requires t7 (true) and wide (true), which means t6 is true.

Going back up the chain:

t5 is true if t6 is true. So t5 is true.

t4 is true if t5 is true. So t4 is true.

t3 is true if t4 is true. So t3 is true.

t2 is true if t3 is true. So t2 is true.

t1 is true if t2 is true. So t1 is true.

t0 is true if t1 is true. So t0 is true.

Finally, q is true if t0 is true. Since t0 is true, q must be true.
</think>

ANSWER: A
\end{Verbatim}

%% file: trace-examples/llama3-8b-sound.tex
\begin{Verbatim}[breaklines=true,breakanywhere=true,fontsize=\scriptsize,frame=single,framesep=3pt,samepage=false]
Let's reason step by step to determine the truth value of proposition q.

From the rules, we can deduce the following:

1. g2_0 and g2_1 are true.
2. x3 is true (from cq being true).
3. x2 is not true (since x3 is true).
4. x1 is true (since x2 is not true).
5. x0 is not true (since x1 is true).
6. t7 is true (since btrue is true).
7. wide is true (from p0 and p1 being true).
8. t6 is true (since t7 and wide are true).
9. t5 is true (since t6 is true).
10. t4 is true (since t5 is true).
11. t3 is true (since t4 is true).
12. t2 is true (since t3 is true).
13. t1 is true (since t2 and cq are true).
14. q is true (since t0 is true, which is true since t1 is true).

Since we have found a set of atoms that satisfies all the rules, we can conclude that proposition q is true in at least one answer set.

Therefore, the answer is:

ANSWER: A
\end{Verbatim}

%% file: trace-examples/llama3-8b-unsound.tex
\begin{Verbatim}[breaklines=true,breakanywhere=true,fontsize=\scriptsize,frame=single,framesep=3pt,samepage=false]
To determine the truth value of proposition q, I will start by tracing its dependencies back to the foundation.

Proposition q is true if proposition t0 is true.
Proposition t0 is true if proposition t1 is true.
Proposition t1 is true if proposition t2 is true.
Proposition t2 is true if proposition t3 is true.
Proposition t3 is true.

Since proposition t3 is already established as true, the chain of dependencies ends here. Therefore, I can conclude that:

Proposition q is true if proposition t0 is true and proposition t1 is true and proposition t2 is true and proposition t3 is true.
Proposition t0 is true if proposition t1 is true.
Proposition t1 is true if proposition t2 is true.
Proposition t2 is true if proposition t3 is true.
Proposition t3 is true.

Since all the dependencies are founded, I can conclude that:

Proposition q is true.

ANSWER: A
\end{Verbatim}

%% file: trace-examples/llama3-8b-none.tex
\begin{Verbatim}[breaklines=true,breakanywhere=true,fontsize=\scriptsize,frame=single,framesep=3pt,samepage=false]
Let's break down the problem step by step:

1. Proposition etrue is true.
2. From rule x0, since etrue is true and x1 is not true, x0 must be true.
3. From rule x3, since etrue is true and x0 is true, x3 must be true.
4. From rule cq, since x3 is true, cq must be true.
5. From rule t4, since t5 is true and cq is true, t4 must be true.
6. From rule t1, since t2 is true, t1 must be true.
7. From rule t0, since t1 is true, t0 must be true.
8. From rule q, since t0 is true, q must be true.

Therefore, proposition q is true.

ANSWER: A
\end{Verbatim}